\pdfoutput=1

\documentclass[11pt]{article}
\usepackage[preprint]{acl}

\usepackage{times}
\usepackage{latexsym}

\usepackage[T1]{fontenc}

\usepackage[utf8]{inputenc}

\usepackage{microtype}

\usepackage{inconsolata}

\usepackage{amsthm}
\usepackage{amsmath}
\usepackage{amssymb}
\usepackage{caption}
\usepackage{subcaption}
\usepackage{enumerate}
\usepackage{multirow}
\usepackage{graphicx} 
\usepackage{url}
\usepackage{booktabs}
\usepackage{colortbl}
\usepackage{algorithm}
\usepackage{algorithmic}
\usepackage{makecell}
\usepackage{tcolorbox}
\usepackage{enumitem}
\usepackage{cleveref} 
\crefname{section}{§}{§§}
\Crefname{section}{§}{§§}
\title{Symbolic Working Memory Enhances Language Models for \\ Complex Rule Application}

\author{Siyuan Wang\textsuperscript{\rm 1}, Zhongyu Wei\textsuperscript{\rm 2}, Yejin Choi\textsuperscript{\rm 3,4}, Xiang Ren\textsuperscript{\rm 1} \\
\textsuperscript{\rm 1}University of Southern California,
\textsuperscript{\rm 2}Fudan University, \\
\textsuperscript{\rm 3}University of Washington,
 \textsuperscript{\rm 4}Allen Institute for Artificial Intelligence \\
\texttt{siyuanwang1997@gmail.com} \\
}

\begin{document}
\maketitle
\begin{abstract}
Large Language Models (LLMs) have shown remarkable reasoning performance but struggle with multi-step deductive reasoning involving a series of rule application steps, especially when rules are presented non-sequentially.
Our preliminary analysis shows that while LLMs excel in single-step rule application, their performance drops significantly in multi-step scenarios due to the challenge in rule grounding. It requires anchoring the applicable rule and supporting facts at each step, amidst multiple input rules, facts, and inferred facts. 
To address this, we propose augmenting LLMs with external working memory and introduce a neurosymbolic framework for rule application. The memory stores facts and rules in both natural language and symbolic forms, enabling precise tracking. Utilizing this memory, our framework iteratively performs symbolic rule grounding and LLM-based rule implementation. The former matches predicates and variables of symbolic rules and facts to ground applicable rules at each step. Experiments indicate our framework's effectiveness in rule application and its robustness across various steps and settings~\footnote{Code and data are available at \url{https://github.com/SiyuanWangw/RuleApplication}.}.
\end{abstract}

\section{Introduction}
Large Language Models (LLMs)~\cite{openai2023gpt4, touvron2023llama, team2023gemini, wei2022chain} have demonstrated impressive performance across diverse reasoning tasks. However, they still face challenges with multi-step deductive reasoning~\cite{creswell2022selection, ling2024deductive, lee2024symba}, where LLMs are provided with a set of facts and logical rules, and need to derive an answer to the query through a sequence of rule application steps. Specifically, each step of rule application requires applying a specific rule to its supporting facts to deduce new conclusions.  
\begin{figure}[!ht]
    \centering
    \begin{minipage}{\columnwidth}
    \centering
    \includegraphics[width=\columnwidth]{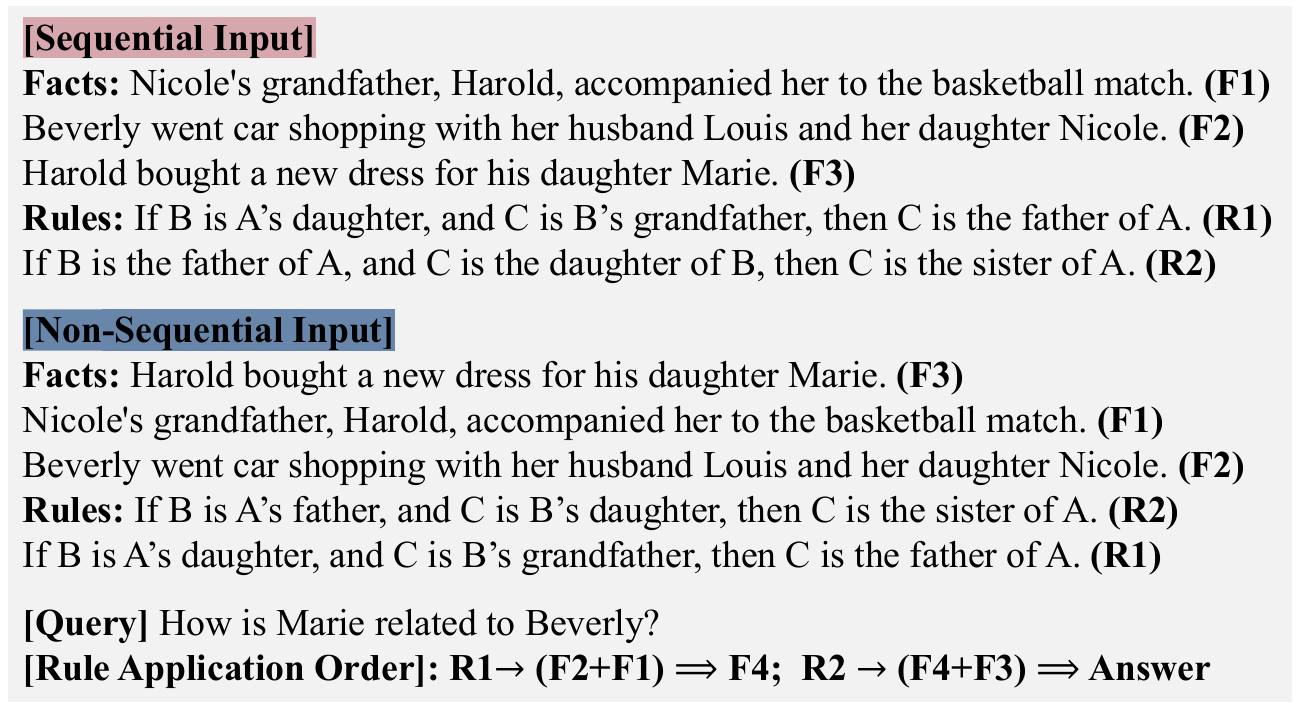}
    \end{minipage}
    \vspace{2.5mm}
    
    \begin{minipage}{\columnwidth}
    \hspace{4mm}
    \includegraphics[width=0.82\columnwidth]{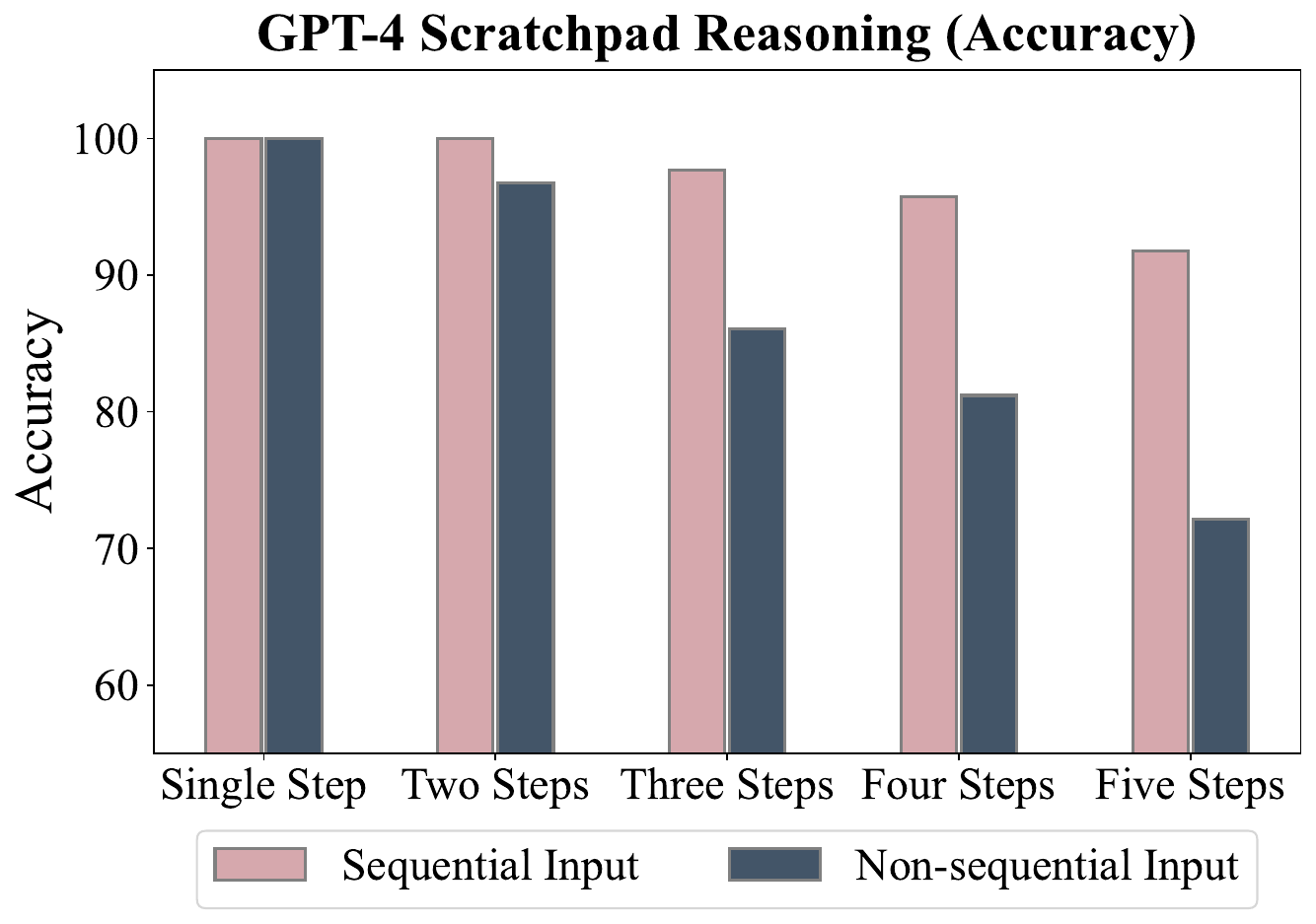}
    \end{minipage}
    \caption{Performance of GPT-4 using scratchpad Chain-of-Thought (CoT) reasoning across various rule application steps on CLUTRR~\cite{sinha2019clutrr}, with an example of two-step rule application shown above.}
    \label{fig:preliminary_analysis}
\end{figure}
Moreover, LLMs especially struggle when the surface patterns deviate from the sequential ordering of the rules~\cite{chen2024premise, berglund2023reversal}. 

We conduct a preliminary analysis of LLM performance across various rule application steps, with rules sequentially and non-sequentially input in their application order. As shown in Figure~\ref{fig:preliminary_analysis}, we observe three phenomena: (1) LLMs are effective at executing single-step rule application. (2) Their performance declines as the number of rule application steps increases. (3) Performance significantly worsens when rules are presented non-sequentially compared to sequentially, especially in long-term reasoning.
Overall, LLMs excel in single-step rule application but face challenges in multi-step rule application, that requires tracking long-term facts and rules and determining appropriate rule and facts for application at each step.

Each step of rule application typically consists of two processes: rule grounding and rule implementation. Rule grounding anchors the current applicable rule with supporting facts from the input, while rule implementation infers new facts based on the identified rule and facts.
The before-mentioned challenges primarily arise from rule grounding using LLMs. Specifically, complex reasoning involves multiple input facts, rules, and intermediate inferred facts, making it difficult to accurately track long-term rule and facts (especially inferred ones) for each step using LLMs' internalized reasoning~\cite{lanchantin2024learning}. 
Additionally, as rules are often provided in a non-sequential order or include irrelevant ones, rule grounding requires referencing back and forth across all rules to identify the applicable one at each step, posing challenges for auto-regressive LLMs~\cite{chen2024premise}.

For precise tracking in multi-step rule application, we propose augmenting LLMs with an external working memory, inspired by humans' extensive use of memory for intelligence tasks~\cite{hardman2016reasoning}. 
It explicitly stores an unlimited list of facts and rules, facilitating easy access during rule grounding, and the writing of new facts after intermediate rule implementation. Besides, it stores rules and facts in a non-ordered manner, minimizing the influence of the input order on LLMs reasoning. We implement this working memory to store rules and facts in both natural language and their symbolic forms (\textit{i.e.}, in Prolog), thus supporting precise symbolic reference.

Building on working memory, we propose a neurosymbolic framework for rule application. This framework uses working memory for symbolic rule grounding and LLMs for rule implementation, leveraging LLMs' effectiveness in single-step rule application. This combination is more flexible than purely symbolic execution and more precise than fully LLM-driven methods. The workflow begins by writing all input facts and rules into working memory. It then proceeds with multiple steps of rule application, each involving symbolic rule grounding followed by LLM-based rule implementation. Specifically, symbolic rule grounding performs predicate and variable matching within the symbolic forms of facts and rules, checking for conflicts to determine the applicable rule with supporting facts.
In rule implementation, LLMs infer new facts based on the grounded rule and facts, and the new inferred facts with their symbolic notations are written into the working memory. This cycle continues until the inferred facts solve the query or a maximum number of steps is reached. 


We conduct experiment on four datasets involving multi-step rule application: CLUTRR and ProofWriter for logical reasoning, AR-LSAT for constraint satisfaction and Boxes for object state tracking. Results show that our framework outperforms CoT-based and symbolic-based baselines using GPT-4 and GPT-3.5, and exhibits robustness across various rule application steps and settings.

\section{Preliminary}
\subsection{Problem Definition}
We consider reasoning tasks involving deductive rule application in natural language, which take a context and a query as input. The context includes all necessary facts and rules for solving the query, 
though they may be non-sequentially provided in their application order and include irrelevant distractors. The model needs to apply specific rules to both the given and intermediate inferred facts to deduce new facts and ultimately output the answer. 

\subsection{External Working Memory}
\begin{figure}[!th]
    \centering
    \includegraphics[width=0.99\columnwidth]{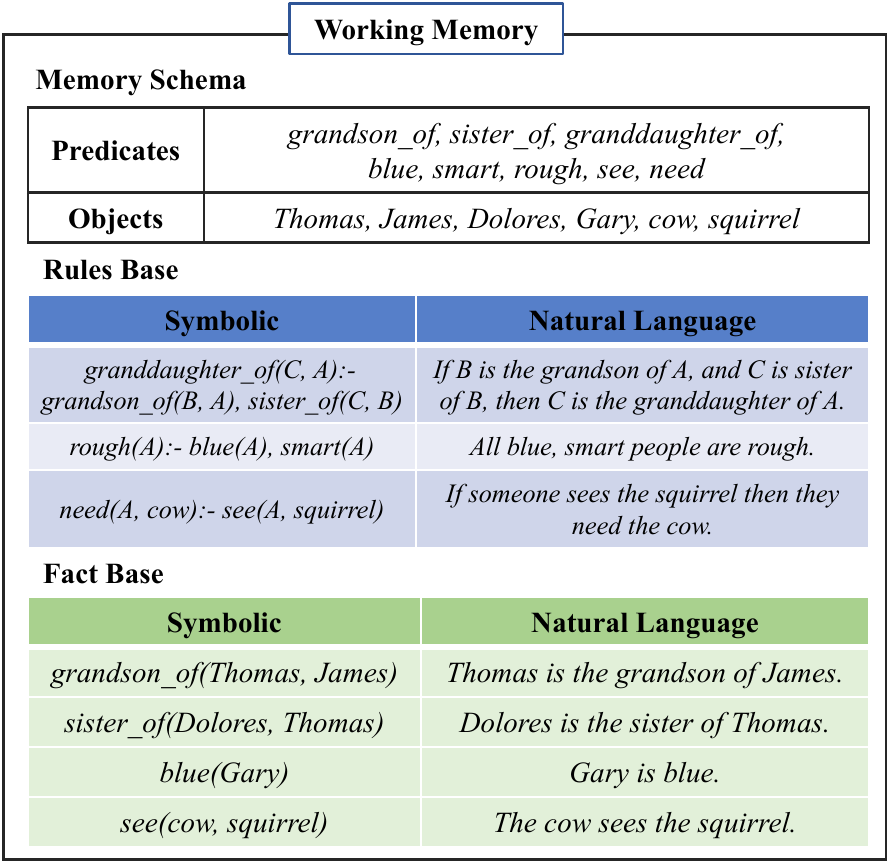}
    \caption{An illustration of the working memory.}
    \label{fig:working_memory}
\end{figure}

To enhance LLMs for precise long-term tracking in multi-step rule application, we introduce an external working memory to explicitly store rules and facts, as illustrated in Figure~\ref{fig:working_memory}.

\vspace{-2mm}
\paragraph{Working Memory Composition}
The working memory consists of three components: a fact base, a rule base and a memory schema. The fact base stores a list of facts from the input context and intermediate reasoning, while the rule base saves a list of input rules. The facts and rules are stored in both natural language and their symbolic forms to support precise symbolic reference and verbalized utilization during multi-step rule application. The memory schema maintains a unified vocabulary of all involved predicates and objects in each instance, avoiding semantic duplication. For example, if ``father\_of'' or ``located\_in'' are in the schema, then ``father-in-law\_of'' or ``located\_at'' will not excluded. The symbolic facts and rules in the memory are constituted using these predicates and objects from the schema.

The working memory supports two operations: read and write. The read operation retrieves necessary facts and rules from the memory. 
The write operation involves adding new rules or facts to the memory, or updating existing facts. 
The decision to add or update facts depends on whether the context involves fact updating, such as an object's location changing over time. If new facts conflict with existing ones, updating occurs; otherwise, new facts are added. In contrast, for static information like the kinship relationship between individuals, new inferred facts will never conflict with existing ones, allowing them to be directly added.


\paragraph{Symbolic Formulation}
Facts and rules are symbolically represented using Prolog notations~\cite{apt1997logic}. Specifically, a fact is a predicate expression with several arguments, formatted as \textit{predicate(arg1, arg2, ...)}, where \textit{args} are specific objects. For example, the fact ``\textit{Dolores is the sister of Thomas.}'' can be formulated as \textit{``sister\_of(Dolores, Thomas)}''. A rule typically takes the form \textit{conclusion:-premises}, interpreted as \textit{If premises, then conclusion.} Both the conclusion and premises are composed of atomic facts, where \textit{args} including both abstract variable symbols like \textit{A, B, C} and specific objects. For example, ``\textit{If B is the grandson of A, and C is sister of B, then C is the granddaughter of A}'' can be represented as \textit{granddaughter\_of(C, A):-grandson\_of(B, A), sister\_of(C, B)}. More examples are in Figure~\ref{fig:working_memory}.

\paragraph{Memory Schema}
A key challenge in managing working memory is ensuring no duplication caused by different expressions conveying the same semantic meaning. This is essential for updating facts and identifying applicable rules based on supporting facts. To address this, we establish a memory schema for maintaining canonical predicates and objects. Symbolic facts and rules are formulated using predicates and objects from this schema.

The schema is dynamically constructed throughout the symbolic formulation process. Initially, the schema is empty. 
When formulating each fact or rule, the process first looks up whether the existing memory schema can accommodate the necessary predicates and objects to encode that piece of information. If it can, symbolic formulation is conducted directly based on the memory schema. If it cannot, new predicates or objects are created and added to the memory schema, and the symbolic formulation proceeds using these additions. The dynamic construction process of the memory schema can be viewed in Appendix~\ref{app:memory_schema}.

\section{Framework}
Complex reasoning often necessitates multi-step rule application amid non-sequential and irrelevant rules and fact. To address this, we propose a two-stage paradigm for each rule application step: rule grounding and rule implementation. Rule grounding anchors the applicable rules and supporting facts at each step. Rule implementation then infers new facts based on the grounded rules and facts. 

\begin{figure*}[!th]
    \centering
    \includegraphics[width=2.0\columnwidth]{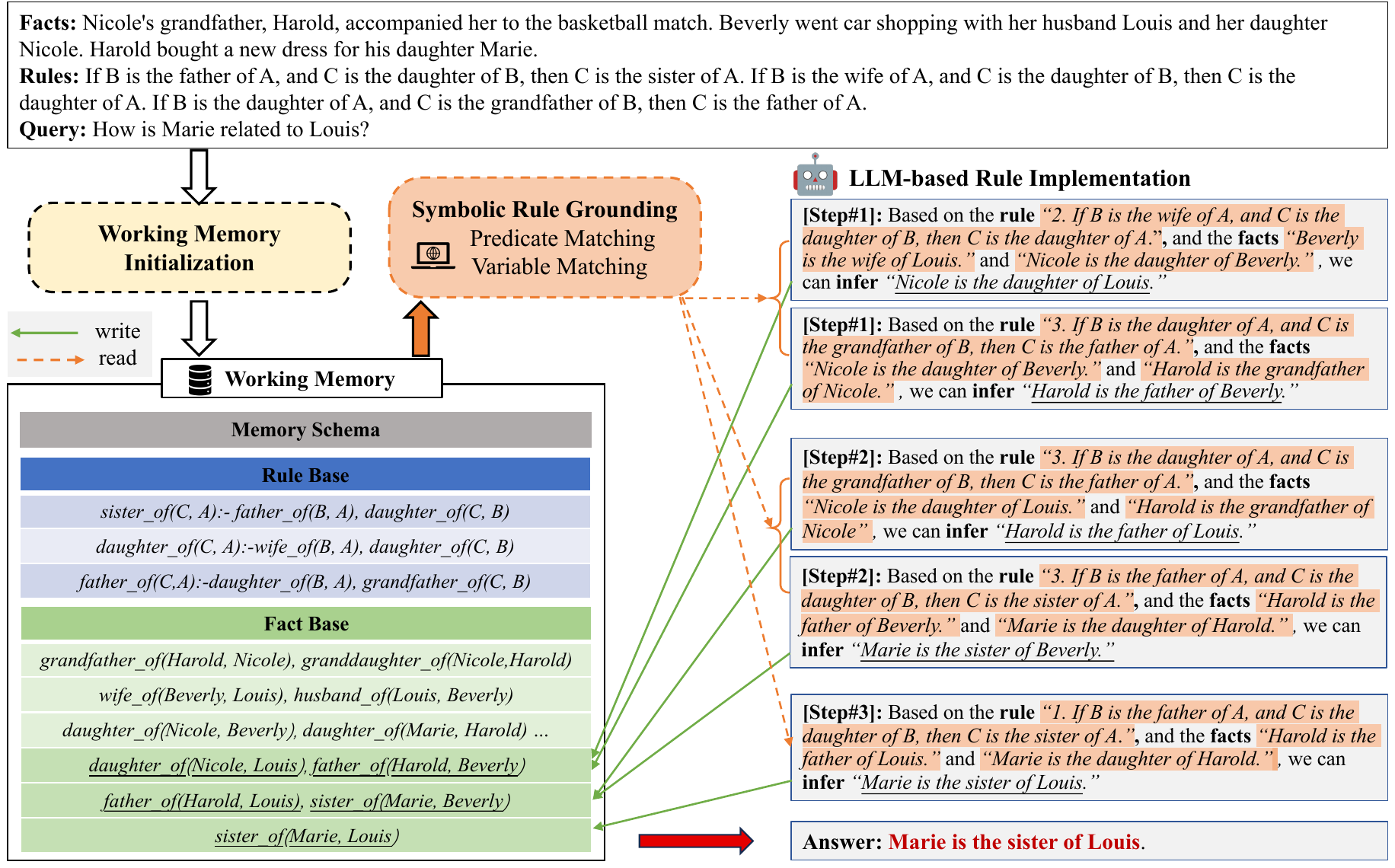}
    \caption{The workflow of our neurosymbolic rule application framework based on working memory. Details of the memory schema and natural language expressions of facts and rules are omitted in the memory for simplicity.}
    \label{fig:framework_workflow}
\end{figure*}

Following this paradigm, we introduce a working memory-based neurosymbolic framework for rule application. It first initializes the working memory with all facts and rules from the input context. It then iteratively performs multi-step rule application, each step involving symbolic rule grounding based on symbolic formulations of facts and rules, followed by LLMs-based rule implementation. This process continues until the query is solved or a maximum number of steps is reached. The detailed workflow is shown in Figure~\ref{fig:framework_workflow}.

\subsection{Working Memory Initialization}
To comprehensively initialize the working memory from the input context, we first decompose the context into multiple sentences. Then we prompt LLMs to list existing facts and rules for each sentence within the context. This involves extracting the natural language expressions and simultaneously parsing their symbolic formulations based on the memory schema. Both the natural language and symbolic representations of all facts and rules are then written into the working memory. Any new predicates and objects beyond the memory schema are also incorporated into the working memory. 
The detailed prompt can be found in Appendix~\ref{app:prompts}.

\subsection{Symbolic Rule Grounding}
\begin{figure}[!th]
    \centering
    \includegraphics[width=1.0\columnwidth]{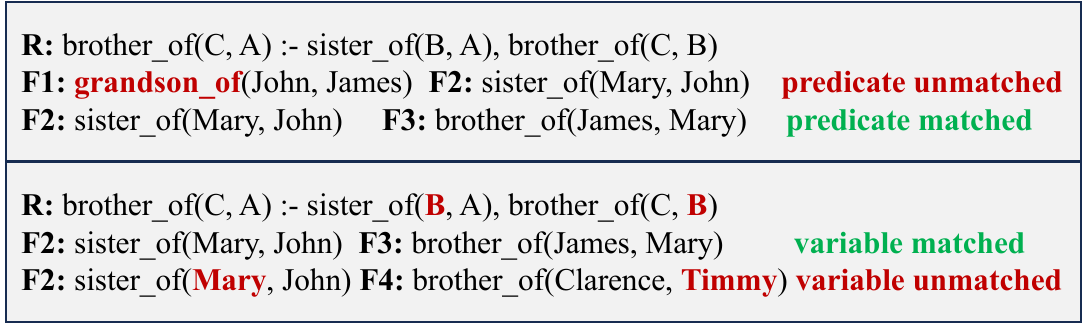}
    \caption{Examples of predicate and variable matching.}
    \label{fig:matching}
\end{figure}

At each step of rule application, we first ground the current applicable rules and corresponding supporting facts from the working memory. We adopt a symbolic predicate and variable matching strategy between facts and rules for precise grounding.
\begin{itemize}[itemsep=1pt, leftmargin=10pt, parsep=2pt, topsep=2pt]
\item \noindent\textbf{Predicate Matching} checks if the predicates of selected facts match those of the rule's premises. This exact string matching can be further relaxed using approximate string or model-based semantic matching to accommodate parsing inconsistencies for more flexible grounding. 
\item \noindent\textbf{Variable Matching} verifies whether the arguments of facts can instantiate the variables in rule premises without conflicts (\textit{i.e.}, each variable is instantiated by the same argument), or can match the objects in rule premises.
\end{itemize}
Detailed examples are illustrated in Figure~\ref{fig:matching}. We observe that the predicates of facts \textit{F1} and \textit{F2} do not match with rule \textit{R}, while the arguments of \textit{F2} and \textit{F4} cannot instantiate the variable \textit{B} in rule \textit{R}. After this symbolic rule grounding, rule \textit{R} is applicable to its supporting facts  \textit{F2} and \textit{F3}.

Specifically, we adopt different rule grounding approaches for various tasks types. For tasks like logical reasoning, where \textbf{facts have no inherent chronological order and a single fact never involves updating}, we adopt exhaustive enumeration for rule grounding. We enumerate all combinations of facts for each rule according to the number of premise facts, and check all rules. We perform both predicate and variable matching, deeming a rule applicable if no conflicts arise with the corresponding facts. 
Notably, each set of supporting facts for the current step's applicable rules must include the newly inferred facts from the previous round to avoid repeating rule implementation.
For particular constraint satisfaction tasks where all rules need to be satisfied with diverse constraint predicates, we only conduct variable matching to rank the most applicable rule at each step. 

For tasks like object state tracking, where \textbf{facts follow an inherent sequential order due to temporal operations}, causing single state facts to update over time, we perform rule grounding according to the chronological order of given operations. For the operational fact at each step, we identify the most applicable rule and relevant state facts based on both predicate matching and variable matching.

Most reasoning tasks that involve rule application can be categorized into two main types: static reasoning and dynamic operational decision-making. These tasks can be approached using above two rule grounding strategies: exhaustive enumeration and chronological grounding.


\subsection{LLM-based Rule Implementation}
LLMs are effective at single-step rule application. After symbolic rule grounding that identifies the applicable rules and corresponding supporting facts from the working memory at each step, we leverage LLMs to implement all applicable rules in parallel. Specifically, we input each rule with its supporting facts and prompt LLMs to infer possible new facts in both natural language and symbolic formulations. The inferred facts are then written into the working memory accordingly. During each step of rule implementation, we also determine whether newly inferred fact solves the query (for logical reasoning) or check for rule-facts conflicts (for constraint satisfaction). 

\paragraph{Final Answer Prediction} If a new fact resolves the query, the iteration ends and we utilize that fact for the final answer. For multi-choice constraint satisfaction, we select the option without conflict as the final answer (or reversely taking the option with conflict for negative questions). For object state tracking where iteration ends only after all operations, the query can be directly answered by looking up the query object's state from the working memory. If all inferred facts in each step cannot solve the query, the process will proceed to the next iteration. The cycle continues until the query is resolved or a maximum step count is reached. If the query remains unsolved, we employ a backup CoT method to output the final answer. Detailed prompts are provided in Appendix~\ref{app:prompts}.

\begin{table*}[!th]
    \centering
    \resizebox{0.96\textwidth}{!}{
    \begin{tabular}{ccccccccc}
    \toprule
    \multirow{2}{*}{Method} & \multicolumn{2}{c}{CLUTRR} & \multicolumn{2}{c}{ProofWriter} & \multicolumn{2}{c}{AR-LSAT} & \multicolumn{2}{c}{Boxes} \\
     & GPT-4 & GPT-3.5 & GPT-4 & GPT-3.5 & GPT-4 & GPT-3.5 & GPT-4 & GPT-3.5 \\
    \midrule
    \rowcolor[HTML]{E1E1E1} \multicolumn{9}{c}{\textit{CoT-base Methods}} \\
    \midrule
    Scratchpad-CoT & 83.83\% & 57.02\% & 61.33\% & 49.67\% & 41.25\% & 30.00\% &91.85\% &15.60\% \\
    SC-CoT & 85.53\% & 59.57\% & 62.00\% & 54.00\% & 45.00\% & 31.25\% & 93.33\% & 17.04\% \\
    Self-Notes & 74.04\% & 55.74\% & 62.00\% & 52.67\% & 47.50\% & 23.75\% & 92.59\% & 18.52\% 
    \\ \midrule
    \rowcolor[HTML]{E1E1E1} \multicolumn{9}{c}{\textit{Symbolic-based Methods}} 
    \\ \midrule
    Logic-LM & / & / & 62.33\% & 52.00\% & 50.00\% & 31.25\% & / & / \\
    SymbCoT & / & / & 65.67\% & 51.33\% & 60.00\% & 21.25\% & / & / \\
    \rowcolor[HTML]{E1E1E1} 
    WM-Neurosymbolic & \bf 92.34\% & \bf 78.72\% & \bf 77.33\% & \bf 58.00\% & \bf 70.00\% & \bf 35.00\% & \bf 100\% & \bf 34.29\% \\
    \bottomrule
    \end{tabular}
    }
    \caption{Experimental results (accuracy \%) of different methods using GPT-4 and GPT-3.5-turbo\protect\footnotemark.}
    \label{overall_result}
\end{table*}
\footnotetext{The results we report of Logic-LM on ProofWriter are lower than the performance stated in its paper. This is because we re-implement it on our sampled subset (reasoning depths 3-5), which is more challenging than the original \textit{depth-5} subset that actually includes reasoning depths from 0 to 5.}
\section{Experiments}
\subsection{Setup}
\paragraph{Datasets} We conduct experiments on four reasoning datasets that involve multi-step of deductive rule application, including CLUTRR~\cite{sinha2019clutrr}, ProofWriter~\cite{tafjord2020proofwriter}, AR-LSAT~\cite{zhong2021ar} and Boxes~\cite{kim2023entity}, detailed as follows:
\begin{itemize}[itemsep=0.5pt, leftmargin=12pt, parsep=1pt, topsep=1pt]
    \item \textbf{CLUTRR and ProofWriter} are two logical reasoning datasets, involving the application of commonsense and predefined logical rules. For CLUTRR, we select 235 test instances requiring 2-6 steps of rule application. For ProofWriter, we select instances necessitating 3-5 of reasoning steps from the open-world assumption subset, totaling 300 instances with balanced labels.
    \item \textbf{AR-LSAT} is a constraint satisfaction dataset sourced from the Law School Admission Test, and requires applying all conditional rules to find satisfactory solutions. Since multiple instances in the original dataset share the same context, which may deviate the evaluation, we select all instances with unique contexts from both the development and test sets, resulting in 80 examples for our evaluation.
    \item \textbf{Boxes} requires reasoning about objects' states after multiple operations, where apply inferential rules for these operations can enhance reasoning. We collect all 135 instances involving 6-7 operations to ensure evaluation difficulty. As rules are not provided, we manually curate the corresponding rule for each operation. 
\end{itemize}

\paragraph{Baseline}
We compare our framework with two types of baselines: CoT-based methods and symbolic-based methods. The CoT-based methods include: (1) Scratchpad-CoT~\cite{nye2021show, wei2022chain} performs chain-of-thought reasoning in a scratchpad manner after the entire input; (2) Self-Consistency CoT (SC-CoT)~\cite{wang2022self} samples three reasoning paths and takes the majority vote as the final predication. Specifically, we shuffle input order for the first three tasks and adopt different temperatures (\textit{i.e.}, 0, 0.5, 1.0) for the last task for sampling; (3) Self-Notes~\cite{lanchantin2024learning} prompts the model to generate multiple internal reasoning notes interleaving with the input.
The symbolic-based methods include: (4) Logic-LM~\cite{pan2023logic} utilizes LLMs to parses natural language problems into symbolic formulations and then performs deterministic inference with symbolic solvers, like Z3 theorem prover~\cite{de2008z3}; and (5) SymbCoT~\cite{xu2024faithful} fully utilizes LLMs to parse language facts and rules into symbolic expressions and solve problems step-by-step by CoT.

Our working memory-based neurosymbolic framework, WM-Neurosymbolic, is implemented based on two different backbone LLMs: GPT-4 (gpt-4-turbo-0409 for CLUTRR, ProofWriter and Boxes, gpt-4o for AR-LSAT) and GPT-3.5 (gpt-3.5-turbo-0125). This enables evaluation of its effectiveness with various abilities of symbolic semantic parsing and one-step rule application.
We adopt one-shot prompting strategy for CoT-based baselines, while symbolic-based methods, which require better output format control in sub-procedures, use few-shot prompts with multiple examples. 
Similarly, WM-Neurosymbolic employs few-shot prompts, but we try to ensure all examples in each prompt belong to a single instance for a fair comparison. We also provide comparisons with multi-shot CoT-based methods in Appendix~\ref{app:more_shot_cot_baseline}, according to the maximum number of examples used by our framework in each dataset.
More implementation details are available in Appendix~\ref{app:implementation}.

\subsection{Overall Performance}
The overall results are presented in Table~\ref{overall_result}. For symbolic-based methods, which may fail to return an answer caused by symbolic formulation errors, we use Scratchpad-CoT as a backup. We have the following observations:
\begin{enumerate}[label=(\arabic*), itemsep=0.5pt, leftmargin=18pt, parsep=0.5pt, topsep=0.5pt]
\item Our method significantly outperforms all baselines across all datasets, including the extremely challenging AR-LSAT dataset, demonstrating the superiority of our working memory-based neurosymbolic framework.
\item Our framework is effective on top of different LLM backbones with varying abilities in symbolic parsing and one-step rule application. Specifically, GPT-3.5-based framework shows significant improvement on formally expressed problems (CLUTRR, Boxes) while GPT-4 excels at more naturalistic problems (ProofWriter, AR-LSAT). This suggests our framework are more effective as backbone LLMs advance.
\item Compared to previous symbolic-based methods that perform both rule grounding and implementation either symbolically or by LLMs, our framework exhibits improvement, demonstrating flexibility and robustness by disentangling rule grounding and implementation, respectively symbolically and through LLMs.
\end{enumerate}

\subsection{Ablation Study}
To investigate the effectiveness of different stages in our framework, we conduct an ablation study taking GPT-4 as the backbone on the CLUTRR and ProofWriter datasets\footnote{To save computational costs, we select instances from ProofWriter that require 5 reasoning steps for analysis.}. We substitute decomposed-based memory initialization with scratchpad-CoT initialization, symbolic rule grounding with LLM-based grounding, and LLM-based rule implementation with symbolic implementation, respectively. Scratchpad-CoT initialization involves formulating all facts and rules within the entire context at once via scratchpad-CoT. LLM-based grounding prompts LLMs to iteratively determine the applicable rules with associated facts at each steps (similar to SELECTION-INFERENCE method~\cite{creswell2022selection}). Symbolic implementation is a deterministic process defined by ourselves.
\begin{table}[!ht]
    \centering
    \setlength\tabcolsep{3.5pt}
    \resizebox{0.49\textwidth}{!}{
    \begin{tabular}{l|cc}
    \toprule
    \multicolumn{1}{c}{Method}  & CLUTRR & ProofWriter \\
    \midrule
    WM-Neurosymbolic & 92.34\% & 74.67\%   \\
    $\rightarrow$ \text{Scratchpad Initialization} & 86.81\% & 66.67\%  \\
    $\rightarrow$ \text{LLM-based Grounding} & 82.98\% & 73.33\%  \\
    $\rightarrow$ \text{Symbolic Implementation} & 90.64\% &  52.00\%   \\
    Scratchpad-CoT & 83.83\% & 53.33\%  \\
    \bottomrule
    \end{tabular}
    }
    \caption{Ablation study based on GPT-4. The arrows denote the replacement of corresponding stages in our framework with specified components.}
    \label{ablation_study}
\end{table}

As shown in Table~\ref{ablation_study}, all substitutions lead to significant performance drops, underscoring the effectiveness of our framework design. Compared to scratchpad-CoT initialization, the decomposed-based strategy simplifies fact and rule formulation by breaking down the context into individual sentences, achieving more comprehensive initialization and improved reasoning. 
LLM-based rule grounding even performs worse than the baseline on CLUTRR, revealing LLMs' deficiency in determining rule application order and tracking long-term facts in multi-step reasoning. However, it shows only a slight drop on ProofWriter, because its reasoning involves a single object, reducing complexity for LLMs. Symbolic implementation causes a greater decline in ProofWriter than in CLUTRR, indicating that advanced LLMs are more robust at one-step rule application for more naturalistic, complex problems than symbolic solvers.

\subsection{Effectiveness on Open-source LLMs} 
To showcase the effectiveness of our framework using affordable open-source LLMs, we implement it on Llama-3-8B-Instruct and compare the results with LLama-based CoT baselines on the CLUTRR and ProofWriter datasets. As shown in Table~\ref{open_source_results}, our framework exhibits robust effectiveness on both closed-source and open-source models. 
\begin{table}[!th]
    \begin{center}
    \resizebox{0.46\textwidth}{!}{
    \begin{tabular}{ccc}
    \toprule
    \multirow{2}{*}{Method}  & CLUTRR  & ProofWriter \\
    & LLama3-8B & LLama3-8B \\
    \midrule
    Scratchpad-CoT & 52.77\% & 50.33\% \\
    SC-CoT & 54.47\% & 53.67\% \\
    Self-Notes & 51.49\% & 52.33\% \\
    \midrule
    WM-Neurosymbolic & 63.40\% & 58.67\% \\
    \bottomrule
    \end{tabular}
    }
    \end{center}
    \caption{Result on LLama-3-8B-Instruct.}
    \label{open_source_results}
\end{table}

\section{Further Analysis}
\begin{figure*}[!th]
    \centering
    \begin{minipage}[b]{0.39\textwidth}
    \includegraphics[width=\columnwidth]{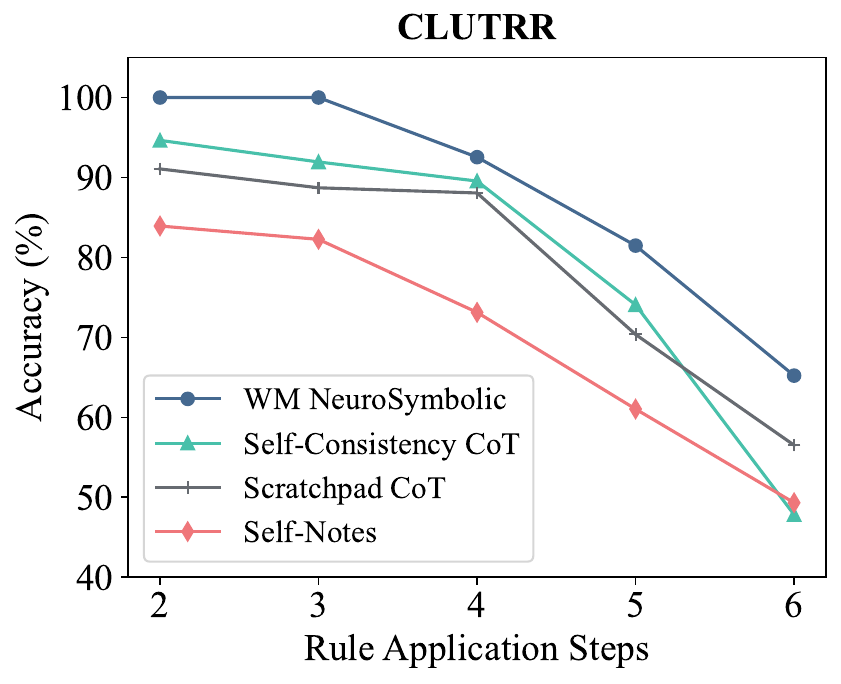}
    \end{minipage}
    \hspace{0.05\textwidth} 
    \begin{minipage}[b]{0.38\textwidth}
    \includegraphics[width=\columnwidth]{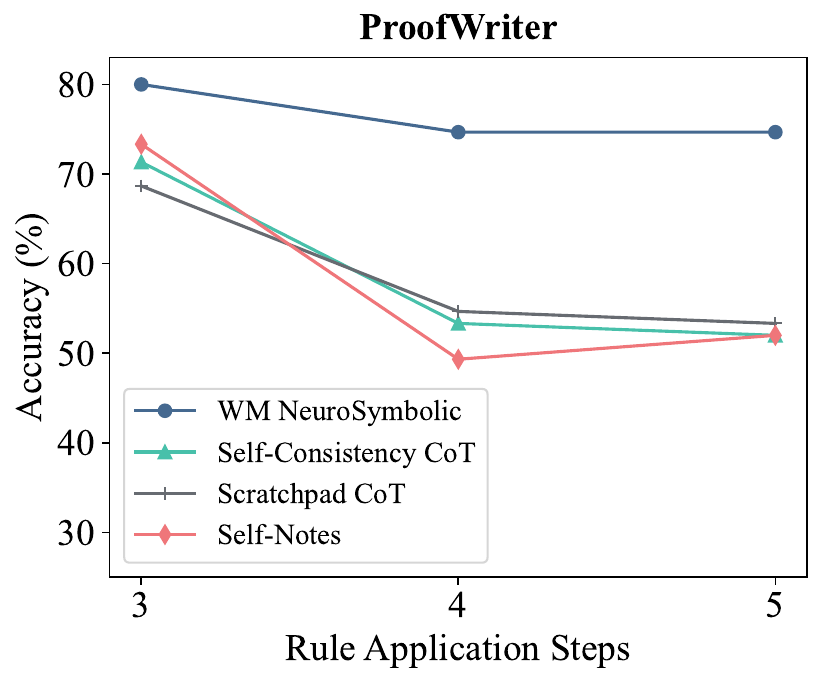}
    \end{minipage}
    \caption{Performance across varying steps of rule application.}  
    \label{fig:depth_analysis}
\end{figure*}

\subsection{Varying Rule Application Steps}
To evaluate the effectiveness of our framework across different steps of rule application, we report the performance of various GPT-4-based methods on the CLUTRR and ProofWriter datasets, which involves 2-6 steps and 3-5 steps. As shown in Figure~\ref{fig:depth_analysis}, our framework consistently performs the best across all steps. As problem complexity increases with more steps, our advantage remains significant. Moreover, Self-Consistency CoT outperforms the baseline CoT on fewer steps, but this advantage diminishes with more steps due to the increased likelihood of generating discrepancies. This can be mitigated by executing more sampling.

\subsection{Different Rule Settings}
In real-world questions, rules are presented in various ways as follows. (1) Ordered Rules: rules are arranged in their application order. (2) Shuffled Rules: rules are provided in a random order. (3) Noisy Rules: rules are shuffled and include irrelevant ones. This setup closely aligns with real-world retrieved-based scenarios where logical rules are retrieved from external sources and may contain distractors. We discuss these three rules settings using the CLUTRR dataset (focusing on 5-6 rule application steps) and compare our framework to CoT-based baselines on GPT-4. Since self-consistency CoT involves shuffling input order, we do not report its performance. For noisy rules, we manually add two irrelevant rules to distract each instance.
\begin{table}[!ht]
    \begin{center}
    \resizebox{0.49\textwidth}{!}{
    \begin{tabular}{c|ccc}
    \toprule
    Rule Settings  & Ordered & Shuffled & Noisy \\
    \midrule
    Scratchpad-CoT & 66\% & 64\% & 58\% \\
    Self-Notes & 68\% & 54\% & 50\% \\
    WM-Neurosymbolic & 74\% & 74\% & 76\% \\
    \bottomrule
    \end{tabular}
    }
    \end{center}
    \caption{Performance on different rule settings.}
    \label{rule_setting}
\end{table}

Table~\ref{rule_setting} shows that CoT-based baselines are susceptible to perturbations from rule order and noise, especially the Self-Notes method. In contrast, our framework exhibits robust effectiveness across all rule settings, even with noisy distractors. Notably, our framework outperforms CoT-based baselines even in the ordered rule setting, underscoring its enhanced ability to precisely track facts at each step and iteratively perform multi-step rule application. Moreover, we implement our framework without rules provided in Appendix~\ref{app:not_rule_provided} to simulate some realistic scenarios where rules are typically well-established commonsense principles derived from real-world observations but not explicitly input.

\subsection{Symbolic Investigation} 
Symbolic-based methods inevitably lead to execution failures due to syntax or semantic errors during symbolic formulation, even performed by an LLM parser. To mitigate this, our framework decouples the symbolic rule application process into executing rule grounding symbolically and rule implementation based on LLMs.
To illustrate our framework's flexibility and efficacy, we report its execution success rate and accuracy across all datasets. Specifically, the execution rate denotes the proportion of instances that can be directly solved by our neurosymbolic framework without backup, and accuracy is calculated for these executable instances.
\begin{table}[!ht]
    \vspace{-1mm}
    \begin{center}
    \setlength\tabcolsep{3pt}
    \resizebox{0.47\textwidth}{!}{
    \begin{tabular}{c|cc|cc}
    \toprule
    \multirow{1}{*}{Executable} & \multicolumn{2}{c|}{GPT-4} & \multicolumn{2}{c}{GPT-3.5} \\
     Statistics & Rate & Accuracy & Rate & Accuracy \\
    \midrule
    CLUTRR & 68.94\% & 100.00\% & 57.02\% & 97.76\% \\
    ProofWriter & 67.00\% & 85.57\% & 67.67\% & 85.22\% \\
    AR-LSAT & 56.25\% & 93.33\% & 12.50\% & 70.00\%\\
    Boxes & 100.00\% & 100.00\% & 100.00\% & 34.29\% \\
    \bottomrule
    \end{tabular}
    }
    \end{center}
    \caption{Execution rate and accuracy statistics for our framework based on GPT-4 and GPT-3.5.}
    \label{execution_statistics}
\end{table}

As depicted in Table~\ref{execution_statistics}, our framework successfully executes over 50\% of instances for all datasets on both GPT-4 and GPT-3.5, except for the complex AR-LSAT dataset on GPT-3.5. Additionally, it achieves high accuracy on executable instances.
In contrast, Logic-LM executes fewer than 10\% of ProofWriter instances, with 33.75\% and 8.75\% of AR-LSAT instances executable based on GPT-4 and GPT-3.5, respectively.\footnote{These figures are obtained from our re-implementation.}
This demonstrates the flexibility of our rule application framework, combining matching-based grounding with LLM-based implementation for a softer symbolic approach. While SymbCoT achieves 100\% execution success, it shows limited accuracy, highlighting the precision of our framework by symbolic grounding.

\subsection{Error Analysis}
We further analyze the cases where our framework incorrectly answers and summarize the major error types. 
(1) Incomplete and inaccurate initialization of the working memory. This primarily occurs when each sentence describes multiple facts or contains coreference, or each instance has inconsistent expressions of predicates with the same meaning even using the memory schema. This issue can be mitigated by utilizing more in-context demonstrations, initializing by sliding every two sentences, or using softer string matching strategies. 
(2) Limited number of LLM-based rule implementation. Since there may be multiple applicable rules at each step, we adopt a pruning method to restrict the maximum numbers of rule implementation at each step to reduce computational costs, making it insufficient to answer some instances. This can be improved by running more rule implementation rounds at each step.
(3) Inaccurate LLM-based rule implementation, especially for challenging tasks like AR-LSAT. This requires employing backbone LLMs with more advanced reasoning capabilities.

\section{Related Work}
\paragraph{LLMs with External Memory}
LLMs~\cite{touvron2023llama, abdin2024phi} have demonstrated remarkable performance across tasks, but struggle with complex reasoning that involves memorizing or grounding long-term information from context or interaction history. 
Beyond extending LLMs' context length~\cite{lee2024human, lu2024longheads}, recent advancements augment LLMs with external memory. 
\citet{park2023generative, guo2023empowering} equip LLMs agents with external memory modules to store and reference long-term dialogue history for better interaction. For knowledge-intensive tasks, ~\citet{yue2024fragrel, wang2024augmenting} encode long-form context into memory for retrieval and utilization. However, previous working memory mainly stores natural language or parametric entries, making accurate referencing and updating challenging. Symbolic memory is further proposed to address this issue.
ChatDB~\cite{hu2023chatdb} uses databases as symbolic memory for precise information recording and processing, but is limited to product inventory. Statler~\cite{yoneda2023statler} introduces symbolic world memory to maintain robot states for embodied reasoning. Our work leverages external memory to store both natural language and symbolic facts and rules, enabling more precise rule grounding for multi-step rule application.

\paragraph{Rule Application for Reasoning}
Rules are well-established principles abstracted from broad real-world observations~\cite{wang2024can, zhu2023large}, or predetermined constraints designed for specific situations~\cite{qiu2023phenomenal}. They serve as a crucial basis for drawing inferences in complex contexts by applying them to known facts to derive new conclusions. 
For example, logical reasoning~\cite{wang2021logic, sun2023indeterminacy, chen2023learning} involves applying rules to contextual facts to answer queries, with~\citet{olausson-etal-2023-linc, pan2023logic} operating in a symbolic manner. Constraint satisfaction~\cite{wang2022lsat} applies rules to find solutions meeting all restrictions. Complex reasoning requires multi-step deductive rule application, each step involving rule grounding and rule implementation for more faithful reasoning~\cite{sanyal2022fairr, creswell2022selection}. We propose to iteratively perform these two processes in a neurosymbolic manner based on working memory.

\section{Conclusion}
In this paper, we augment LLMs with an external working memory and propose a neurosymbolic framework for multi-step rule application to enhance LLMs' reasoning capabilities. The memory stores facts and rules in both natural language and symbolic forms, facilitating accurate retrieval during rule application.
After writing all input facts and rules into the working memory, the framework iteratively performs symbolic rule grounding based on predicate and variable matching, followed by LLM-based rule implementation. It effectively combines the strengths of both symbolic and LLM methods. Our experiments demonstrate the framework's superiority over CoT-based and symbolic-based baselines, and show its robustness across various rule application steps and settings. In the future, we will extend our framework to incorporate more backbone LLMs and datasets, especially on more complex and long-term reasoning tasks.

\section*{Limitations}
\paragraph{Limitation on Experimented Datasets}
Due to computational costs, our work mainly experiments with four datasets, focusing on 
logical reasoning, constraint satisfaction and object state tracking tasks. Future work will include a broader range of tasks and datasets to further validate our framework's effectiveness.

\paragraph{Limitation on Backbone LLMs}
We build our framework upon GPT-4 and GPT-3.5 to demonstrate its effectiveness with various abilities of symbolic semantic parsing and one-step rule application. We will  expand our scope to take more backbone LLMs, including open-source models.

\paragraph{Risk of Environmental Impact}
A significant risk associated with our framework is the potential increase in computational costs and environmental burden due to the extensive use of LLMs APIs. This impact can be mitigated by adopting advanced open-source models like Llama-3-70B that are more efficient with less environmental impact.


\bibliography{custom}

\newpage
\appendix

\section{Memory Schema Update}
\label{app:memory_schema}
\begin{figure*}[!th]
    \centering
    \includegraphics[width=2.0\columnwidth]{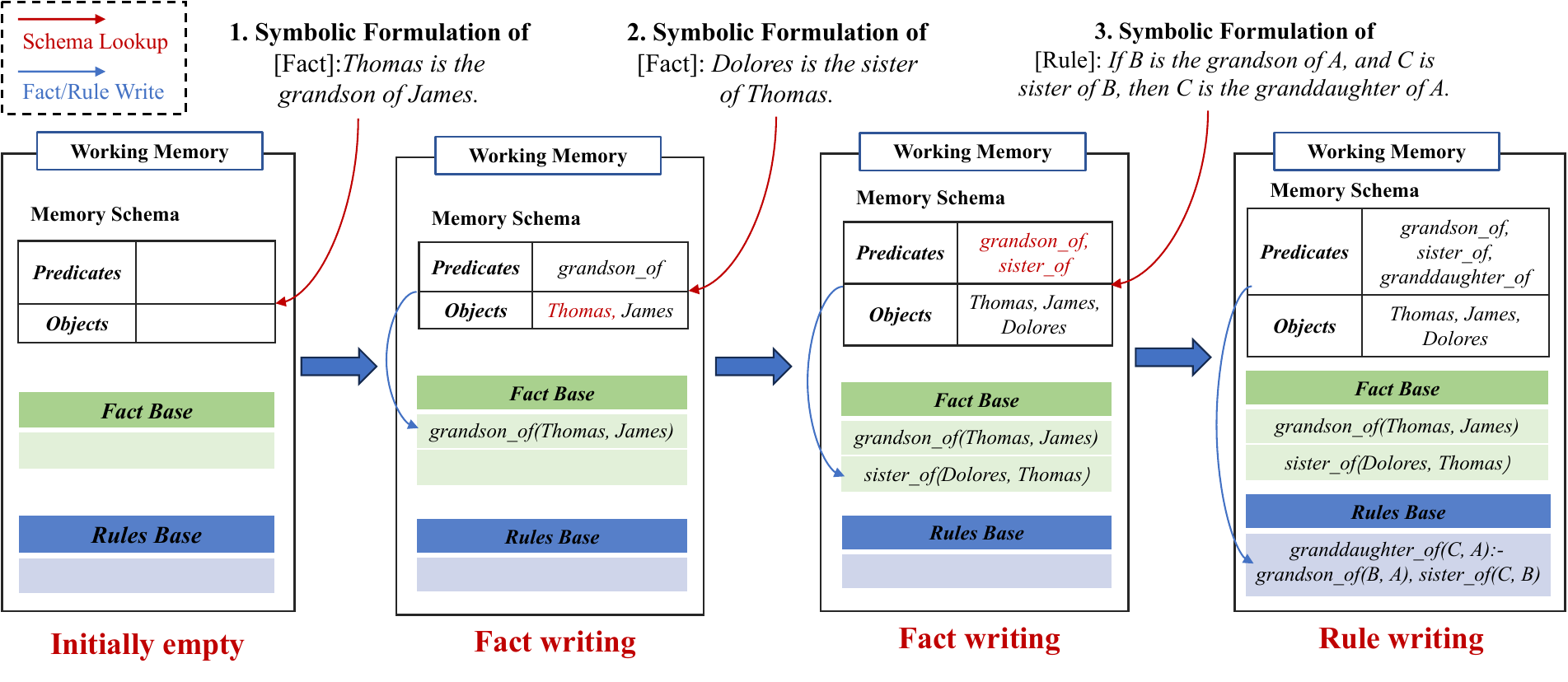}
    \caption{An example construction process of our working memory schema alongside the memory initialization.}
    \label{fig:memory_schema_update}
\end{figure*}

An example of the memory schema construction process is illustrated in Figure~\ref{fig:memory_schema_update}. Before each symbolic formulation, the process first looks up the memory schema to determine whether its maintained predicates and objects can cover the current fact or rule to be formulated. If it can, symbolic formulation is conducted directly based on the memory schema. If it cannot, new predicates or objects are created and added to the memory schema, and the symbolic formulation proceeds based on the updated memory schema. Then new formulated facts and rules are written into the working memory. 

\section{Implementation Details} 
\label{app:implementation}
We implement our framework based on two different backbone LLMs: GPT-4 (gpt-4-turbo-0409 for CLUTRR, ProofWriter and Boxes, gpt-4o for AR-LSAT) and GPT-3.5 (gpt-3.5-turbo-0125), to test its effectiveness with different capabilities of symbolic semantic parsing and one-step rule application. For fair comparison, we re-implement all baseline methods using corresponding LLMs. All CoT-based baselines utilize the same in-context demonstrations. The generation temperature is set to 0.0 by default. The maximum number of steps in our framework is set to 4, 6, 8 for actual 2, 3-4, and 5-6 steps in CLUTRR and ProofWriter. For AR-LSAT, the maximum steps are set according to the number of rules, and for Boxes, they are set according to the number of operational facts.

\section{Further Experiments}
\subsection{Comparison with Multi-shot CoT-based Methods.}
\label{app:more_shot_cot_baseline}
Since we implement WM-Neurosymbolic using few-shot prompts to better control output formats, we conduct additional experiments to illustrate our framework's effectiveness even when compared to CoT-based methods with multi-shot demonstrations. Specifically, we set the number of demonstrations in CoT-based methods for each dataset according to the maximum number of examples used by our framework: 2 for CLUTRR and AR-LSAT, and 3 for ProofWriter and Boxes.
\begin{table*}[!th]
    \centering
    \resizebox{0.96\textwidth}{!}{
    \begin{tabular}{ccccccccc}
    \toprule
     \multirow{2}{*}{Method}  & \multicolumn{2}{c}{CLUTRR} & \multicolumn{2}{c}{ProofWriter} & \multicolumn{2}{c}{AR-LSAT} & \multicolumn{2}{c}{Boxes} \\
     & GPT-4 & GPT-3.5 & GPT-4 & GPT-3.5 & GPT-4 & GPT-3.5 & GPT-4 & GPT-3.5 \\
    \midrule
    \rowcolor[HTML]{E1E1E1} \multicolumn{9}{c}{\textit{One-shot CoT-base Methods}} \\
    \midrule
    Scratchpad-CoT & 83.83\% & 57.02\% & 61.33\% & 49.67\% & 41.25\% & 30.00\% &91.85\% &15.60\% \\
    SC-CoT & 85.53\% & 59.57\% & 62.00\% & 54.00\% & 45.00\% & 31.25\% & 93.33\% & 17.04\% \\
    Self-Notes & 74.04\% & 55.74\% & 62.00\% & 52.67\% & 47.50\% & 23.75\% & 92.59\% & 18.52\% \\
    \midrule
    \rowcolor[HTML]{E1E1E1} \multicolumn{9}{c}{\textit{Multi-shot CoT-base Methods}} \\
    \rowcolor[HTML]{E1E1E1} \textit{Shot Number}  & \multicolumn{2}{c}{2-shot} & \multicolumn{2}{c}{3-shot} & \multicolumn{2}{c}{2-shot} & \multicolumn{2}{c}{3-shot} \\
    \midrule
    Scratchpad-CoT & 86.38\% & 59.57\% & 64.33\% & 48.00\% & 52.50\% & 17.50\% & 97.04\%	& 22.22\% \\
    SC-CoT & 87.23\% & 60.85\% & 66.33\% & 48.33\% & 50.00\% & 18.75\% & 97.78\% & 24.44\% \\
    Self-Notes & 72.76\% & 54.89\% & 61.67\% & 56.33\% & 53.75\% & 21.25\% & 97.04\% & 25.19\% \\
    \midrule
    \rowcolor[HTML]{E1E1E1} WM-Neurosymbolic & \bf 92.34\% & \bf 78.72\% & \bf 77.33\% & \bf 58.00\% & \bf 70.00\% & \bf 35.00\% & \bf 100\% & \bf 34.29\% \\
    \bottomrule
    \end{tabular}
    }
    \caption{Comparison to multi-shot CoT-based methods.}
    \label{more_shot_comparison}
\end{table*}
As shown in Table~\ref{more_shot_comparison}, using more examples in few-shot CoT prompting does not always lead to performance improvement. However, compared to both one-shot and multi-shot CoT-based methods, our framework consistently exhibits enhanced performance.

\subsection{Rule Application without Rules Provided}
\label{app:not_rule_provided}
To simulate realistic scenarios where rules are commonsense principles derived from real-world observations but not explicitly provided, we additionally experiment our framework on CLUTRR and Boxes datasets with rules not pre-defined. Here, our working memory only stores and updates facts. In each step, we select applicable facts (those with overlapping objects) from memory, and ask LLMs to self-generate applicable rules for rule implementation until the query is resolved. 
As shown in Table~\ref{rule_not_given}, compared to the Scratchpad-CoT baseline without provided rules, our framework on top of GPT-4 still shows improvement.
\begin{table}[!ht]
    \begin{center}
    \resizebox{0.45\textwidth}{!}{
    \begin{tabular}{c|cc}
    \toprule
    Methods  & CLUTRR & Boxes \\
    \midrule
    Scratchpad-CoT & 82.13\% & 89.63\% \\
    WM-Neurosymbolic & 83.83\% & 96.30\% \\
    \bottomrule
    \end{tabular}
    }
    \end{center}
    \caption{Performance without rule provided.}
    \label{rule_not_given}
\end{table}

\section{Framework Prompts}
\label{app:prompts}
Table~\ref{prompt_fact_initialization},~\ref{prompt_rule_initialization} and~\ref{prompt_rule_implementation} show the example prompts for fact initialization, rule initialization, and LLM-based rule implementation in the CLUTRR dataset.
Table~\ref{prompt_fact_initialization_ProofWriter},~\ref{prompt_rule_initialization_ProofWriter} and~\ref{prompt_rule_implementation_ProofWriter} show the example prompts for the  ProofWriter dataset. Table~\ref{prompt_fact_initialization_ar_lsat},~\ref{prompt_rule_initialization_ar_lsat} and~\ref{prompt_rule_implementation_ar_lsat} show the example prompts for the  AR-LSAT dataset. Table~\ref{prompt_fact_initialization_boxes},~\ref{prompt_op_fact_initialization_boxes} and~\ref{prompt_rule_implementation_boxes} show the example prompts for the Boxes dataset. 
\begin{table*}
\centering
\resizebox{0.98\textwidth}{!}{
\begin{tcolorbox}[colback=blue!5!white,colframe=black!55!black,width=0.98\textwidth,title={Prompt for Fact Initialization (CLUTRR)}]
\small
Please list all explicitly mentioned facts from the context. \\
Each fact should be presented on a separate line under the header "Facts:". Format each fact as "Person A is the Relationship of Person B." and follow it with its symbolic triplet formatted as "[predicate(A, B)]". \\
For each fact, also provide the corresponding reverse fact. For example, if the fact is "Person A is the Relationship of Person B," the reverse fact is "Person B is the Reverse\_Relationship of Person A. \\
Please try to use the objects and predicates in the provided schema to describe symbolic facts. If the schema does not contain corresponding elements to describe the context, generate the symbolic fact directly from its natural language form. Note: Avoid using objects and predicates that do not exist in the given context when generating facts. \\

\#\#\# Examples: \\
Schema Objects: null \\
Schema Predicates: null \\
Context: Don's father, Joshua, and grandfather, James, went hiking during the first weekend of spring. \\
Facts: \\
- Joshua is the father of Don. [father\_of(Joshua, Don)] \\
- Don is the son of Joshua. [son\_of(Don, Joshua)] \\
- James is the grandfather of Don. [grandfather\_of(James, Don)] \\
- Don is the grandson of James. [grandson\_of(Don, James)] \\
------ \\
Schema Objects: Joshua, Don, James \\
Schema Predicates: father\_of, son\_of, grandfather\_of, grandson\_of \\
Context: James took his daughter Lena out for dinner. \\
Facts: \\
- Lena is the daughter of James. [daughter\_of(Lena, James)] \\
- James is the father of Lena. [father\_of(James, Lena)] \\

\#\#\# Here's what you need to do. \\
\textbf{Schema Objects:} \{objects\}  \\
\textbf{Schema Predicates:} \{predicates\}\\
\textbf{Context:} \{context\} \\
\textbf{Facts:}
\end{tcolorbox}
}
\caption{The prompt for fact initialization in CLUTRR.}
\label{prompt_fact_initialization}
\end{table*}

\begin{table*}
\centering
\resizebox{0.98\textwidth}{!}{
\begin{tcolorbox}[colback=blue!5!white,colframe=black!55!black,width=0.98\textwidth,title={Prompt for Rule Initialization (CLUTRR)}]
\small
Please convert the following inference rule into a symbolic representation in Prolog without changing its wordings. Ensure the conclusion and the premises are separated by ":-". \\
The predicates for each atom should be represented as relationships in lowercase. \\
Please try to use the objects and predicates in the provided schema to describe the symbolic rule. If the schema does not contain corresponding elements to describe the rule, generate the symbolic rule directly from its natural language form. \\

\#\#\# Examples: \\
Schema Objects: Joshua, Don, James \\
Schema Predicates: sister\_of, brother\_of \\
Rule: If B is the sister of A, and C is the brother of B, then C is the brother of A. \\
Symbolic Rule: brother\_of(C, A) :- sister\_of(B, A), brother\_of(C, B). \\
------ \\
Schema Objects: Joshua, Don, James \\
Schema Predicates: sister\_of, father\_of, brother\_of \\
Rule: If B is the father of A, and C is the daughter of B, then C is the sister of A. \\
Symbolic Rule: sister\_of(C, A) :- father\_of(B, A), daughter\_of(C, B). \\

\#\#\# Here's what you need to do. \\
\textbf{Schema Objects:} \{objects\}  \\
\textbf{Schema Predicates:} \{predicates\}\\
\textbf{Rule:} \{rule\} \\
\textbf{Symbolic Rule:}
\end{tcolorbox}
}
\caption{The prompt for rule initialization in CLUTRR.}
\label{prompt_rule_initialization}
\end{table*}

\begin{table*}
\begin{tcolorbox}[colback=blue!5!white,colframe=black!55!black,width=0.98\textwidth,title={Prompt for Rule Implementation (CLUTRR)}]
\small
\textbf{System:} You are an expert in determining kinship relationships. You will receive a query about the kinship between two individuals, and your task is to answer this query.\\

\textbf{User:} At each turn, you will be provided a list of identified supporting facts and an inference rule. \\
Please on a new line starting with "Rule Implementation:" to implement the rule based on the supporting facts to analyze and deduce new potential fact. \\
Then on a new line starting with "New fact:" to outline the new inferred fact in both natural language form and its corresponding symbolic format within "[" and "]". \\
Please try to use the objects and predicates in the provided schema to describe symbolic facts. If the schema does not contain corresponding elements, generate the symbolic fact directly from its natural language form. \\
Finally predict "Yes" or "No" to judge whether the new inferred fact can solve the query, in a new line starting with "Judgement:". \\

\#\#\# Examples: \\
Schema Objects: Joshua, Don, James \\ 
Schema Predicates: sister\_of, brother\_of \\
Query: How is Irvin related to Hugh? \\
Fact List: 3. Frances is the mother of Wesley. 6. Hugh is the son of Frances. \\
Rule: If B is the mother of A, and C is the son of B, then C is the brother of A. \\
Rule Implementation: According to the rule, since Frances is the mother of Wesley, and Hugh is the son of Frances, we can infer that Hugh is the brother of Wesley. \\
New fact: Hugh is the brother of Wesley. [brother\_of(Hugh, Wesley)] \\ 
Judgement: No. Because the new fact does not state the relationship between Irvin and Hugh. \\
------ \\
Schema Objects: Joshua, Leno, James \\ 
Schema Predicates: father\_of, sister\_of, daughter\_of \\
Query: How is Joshua related to Lena? \\
Fact List: 1. James is the father of Joshua. 3. Leno is the daughter of James. \\
Rule: If B is the father of A, and C is the daughter of B, then C is the sister of A. \\
Rule Implementation: According to the rule, since James is the father of Joshua, and Lena is the daughter of James, we can infer that Lena is the sister of Joshua. \\
New fact: Lena is the sister of Joshua. [sister\_of(Lena, Joshua)] \\
Judgement: Yes. Because the new fact states the relationship between Joshua and Lena. \\

\#\#\# Here's what you need to do. \\
\textbf{Schema Objects:} \{objects\} \\
\textbf{Schema Predicates:} \{predicates\}\\
\textbf{Query:} \{query\}\\
\textbf{Fact List:} \{facts\} \\
\textbf{Rule:} \{rule\} \\
\textbf{Rule Implementation:}
\end{tcolorbox}
\caption{The prompt for LLM-based rule implementation in CLUTRR.}
\label{prompt_rule_implementation}
\end{table*}

\begin{table*}
\centering
\resizebox{0.98\textwidth}{!}{
\begin{tcolorbox}[colback=blue!5!white,colframe=black!55!black,width=0.98\textwidth,title={Prompt for Fact Initialization (ProofWriter)}]
\small
Please list the symbolic fact of the given context. \\
Format each symbolic fact in Prolog notation as "predicate(X, Y, ...)" where X, Y, ... are the arguments of the predicate. Avoid predicate nesting such as not(smart(X)), but using not\_smart(X) instead. \\
Please try to use the objects and predicates in the provided schema to describe symbolic facts. If the schema is null or does not contain corresponding elements to describe the context, generate the symbolic fact directly from its natural language form.\\

\#\#\# Examples: \\
Schema Objects: David \\
Schema Predicates: kind \\
Context: Context: Bob is big.  \\
Fact: big(Bob) \\
------ \\
Schema Objects: bald eagle \\
Schema Predicates: needs \\
Context: The cow visits the bald eagle. \\
Fact: visits(cow, bald eagle) \\
------ \\
Schema Objects: lion, squirrel \\
Schema Predicates: sees  \\
Context: The lion does not see the squirrel. \\
Fact: not\_see(lion, squirrel) \\

\#\#\# Here's what you need to do. \\
\textbf{Schema Objects:} \{objects\}  \\
\textbf{Schema Predicates:} \{predicates\}\\
\textbf{Context:} \{context\} \\
\textbf{Fact:}
\end{tcolorbox}
}
\caption{The prompt for fact initialization in ProofWriter.}
\label{prompt_fact_initialization_ProofWriter}
\end{table*}

\begin{table*}
\centering
\resizebox{0.98\textwidth}{!}{
\begin{tcolorbox}[colback=blue!5!white,colframe=black!55!black,width=0.98\textwidth,title={Prompt for Rule Initialization (ProofWriter)}]
\small
Please convert the explicitly provided rule into their symbolic forms in Prolog without changing its original wordings. \\
Format each symbolic rule in Prolog notation with the conclusion and premises separated by ":-", and format each atom fact in the rule as "predicate(X, Y, ...)" where X, Y, ... are the arguments of the predicate. Avoid predicate nesting such as not(smart(X)), but using not\_smart(X) instead. \\
Please try to use the objects and predicates in the provided schema to describe the symbolic rule. If the schema is null or does not contain corresponding elements to describe the rule, generate the symbolic rule directly from its natural language form. Note: Avoid using objects and predicates that do not exist in the provided rule when generating its symbolic form. \\

\#\#\# Examples: \\
Schema Objects: Bob \\
Schema Predicates: kind, smart \\
Rule: If something is kind and smart then it is nice. \\
Symbolic Rule: nice(X) :- kind(X), smart(X) \\
------ \\
Schema Objects: bald eagle \\
Schema Predicates: needs, sees \\
Rule: If someone needs the tiger then the tiger sees the bald eagle. \\
Symbolic Rule: sees(tiger, bald eagle) :- needs(X, tiger) \\
------ \\
Schema Objects: Bob \\
Schema Predicates: kind, big, furry \\
Rule: Kind, big people are not furry. \\
Symbolic Rule: not\_furry(X) :- kind(X), big(X) \\

\#\#\# Here's what you need to do. \\
\textbf{Schema Objects:} \{objects\}  \\
\textbf{Schema Predicates:} \{predicates\}\\
\textbf{Rule:} \{rule\} \\
\textbf{Symbolic Rule:}
\end{tcolorbox}
}
\caption{The prompt for rule initialization in ProofWriter.}
\label{prompt_rule_initialization_ProofWriter}
\end{table*}

\begin{table*}
\begin{tcolorbox}[colback=blue!5!white,colframe=black!55!black,width=0.98\textwidth,title={Prompt for Rule Implementation (ProofWriter)}]
\small
\textbf{System:} You are an expert in logical reasoning. You will receive a context including a list of facts and inference rules, and a specific query. Your task is to answer this query following the provided rule. \\

\textbf{User:} At each turn, you will be provided a list of identified supporting facts and an inference rule. \\
Please on a new line starting with "Rule Implementation:" to implement the rule based on the supporting facts to analyze and deduce new potential fact. \\
Then on a new line starting with "New fact:" to outline the new inferred fact in both natural language form and its corresponding symbolic format within "[" and "]". \\
Please try to use the objects and predicates in the provided schema to describe symbolic facts. If the schema is null or does not contain corresponding elements, generate the symbolic fact directly from its natural language form. \\
Finally predict "Yes" or "No" to judge whether the new inferred fact can solve the query, in a new line starting with "Judgement:". \\

\#\#\# Examples: \\
Schema Objects: Gary \\
Schema Predicates: big, not\_green \\
Query: Is it true that Gary is not red? \\
Fact List: 3. Gary is big. \\
Rule: All big things are not green. \\
Rule Implementation: According to the rule, since Gary is big, we can infer that Gary is not green. \\
New fact: Gary is not green. [not\_green(Gary)] \\
Judgement: No. Because the new fact does not state the relationship between Gary and red. \\
------ \\
Schema Objects: Bob \\
Schema Predicates: furry, big, not\_quiet \\
Query: Is it true that Bob is not quiet? \\
Fact List: 1. Bob is furry. 2. Bob is big. \\
Rule: If Bob is furry and Bob is big then Bob is not quiet. \\
Rule Implementation: According to the rule, since Bob is furry and Bob is big, we can infer that Bob is not quiet. \\
New fact: Bob is not quiet. [not\_quiet(Bob)] \\
Judgement: Yes. Because the new fact states the relationship between Bob and quiet. \\

\#\#\# Here's what you need to do. \\
\textbf{Schema Objects:} \{objects\} \\
\textbf{Schema Predicates:} \{predicates\}\\
\textbf{Query:} \{query\}\\
\textbf{Fact List:} \{facts\} \\
\textbf{Rule:} \{rule\} \\
\textbf{Rule Implementation:}
\end{tcolorbox}
\caption{The prompt for LLM-based rule implementation in ProofWriter.}
\label{prompt_rule_implementation_ProofWriter}
\end{table*}

\begin{table*}
\centering
\resizebox{0.98\textwidth}{!}{
\begin{tcolorbox}[colback=blue!5!white,colframe=black!55!black,width=0.98\textwidth,title={Prompt for Fact Initialization (AR-LSAT)}]
\small
You will receive a context including a list of constraint rules, and a specific query with five candidate options (A, B, C, D, E). \\
Please list the symbolic forms of all established facts in the given query and option. \\
Format each symbolic fact in Prolog notation as "predicate(X, Y, ...)" where X, Y, ... are the arguments of the predicate. \\
Please try to use the objects and predicates in the provided schema to describe symbolic facts. If the schema is null or does not contain corresponding elements to describe the context, generate the symbolic fact directly from its natural language form. Please always use one predicate, i.e., assign. \\

\#\#\# Examples: \\
Schema Objects: Monday, Tuesday, Wednesday, morning \\
Schema Predicates: assign \\
Context: Of the eight students-George, Helen, Irving, Kyle, Lenore, Nina, Olivia, and Robert-in a seminar, exactly six will give individual oral reports during three consecutive days-Monday, Tuesday, and Wednesday. Exactly two reports will be given each day-one in the morning and one in the afternoon-according to the following conditions. \\
Query: If Kyle and Lenore do not give reports, then the morning reports on Monday, Tuesday, and Wednesday, respectively, could be given by \\
Option: A) Helen, George, and Nina \\
Facts: \\
- Helen gives report on Monday morning. [assign(Helen, Monday, morning)] \\
- George gives report on Tuesday morning. [assign(George, Tuesday, morning)] \\
- Nina gives report on Wednesday morning. [assign(Nina, Wednesday, morning)] \\
------ \\
Schema Objects: Monday, Tuesday, Wednesday, morning \\
Schema Predicates: assign \\
Context: Each of seven candidates for the position of judge\u2014Hamadi, Jefferson, Kurtz, Li, McDonnell, Ortiz, and Perkins\u2014will be appointed to an open position on one of two courts\u2014the appellate court or the trial court. There are three open positions on the appellate court and six open positions on the trial court, but not all of them will be filled at this time. The judicial appointments will conform to the following conditions. \\
Query: Which one of the following is an acceptable set of appointments of candidates to courts? \\
Option: E) appellate: Li, Perkins;  trial: Hamadi, Jefferson, Kurtz, McDonnell, Ortiz \\
Facts:  \\
- The appellate court appoints Li and Perkins. [assign(appellate, Li, Perkins)] \\
- The trial court appoints Hamadi, Jefferson, Kurtz, McDonnell and Ortiz. [assign(trial, Hamadi, Jefferson, Kurtz, McDonnell, Ortiz)] \\

\#\#\# Here's what you need to do. \\
\textbf{Schema Objects:} \{objects\}  \\
\textbf{Schema Predicates:} \{predicates\}\\
\textbf{Context:} \{context\} \\
\textbf{Query:} \{query\} \\
\textbf{Option:} \{option\} \\
\textbf{Facts:}
\end{tcolorbox}
}
\caption{The prompt for fact initialization in AR-LSAT.}
\label{prompt_fact_initialization_ar_lsat}
\end{table*}

\begin{table*}
\centering
\resizebox{0.98\textwidth}{!}{
\begin{tcolorbox}[colback=blue!5!white,colframe=black!55!black,width=0.98\textwidth,title={Prompt for Rule Initialization (AR-LSAT)}]
\small
Please list the symbolic forms of the given constraint rule in Prolog without changing its original wordings. \\
Format each symbolic rule in Prolog notation, representing it either as a conclusion or as a combination of a conclusion and premises, separated by ":-". Format each atom fact in the rule as "predicate(X, Y, ...)" where X, Y, ... are the arguments of the predicate. Avoid predicate nesting such as not(smart(X)), but using not\_smart(X) instead. Avoid mathematic expression such as N =< 4, but using samller\_than(N, 4). \\
Please try to use the objects and predicates in the provided schema to describe the symbolic rule. If the schema does not contain corresponding elements, generate the symbolic rule directly from its natural language form. Please always use one predicate, i.e., constraint. \\

\#\#\# Examples: \\
Schema Objects: Monday, Tuesday, Wednesday, morning, Kyle, Lenore, Helen, George, Nina \\
Schema Predicates: constraint \\
Context: Of the eight students-George, Helen, Irving, Kyle, Lenore, Nina, Olivia, and Robert-in a seminar, exactly six will give individual oral reports during three consecutive days-Monday, Tuesday, and Wednesday. Exactly two reports will be given each day-one in the morning and one in the afternoon-according to the following conditions. \\
Constraint Rule: Tuesday is the only day on which George can give a report. \\
Symbolic Rule: \\
- constraint(George, Tuesday) \\
------ \\
Schema Objects: Monday, Tuesday, Wednesday, morning, Kyle, Lenore, Helen, George, Nina \\
Schema Predicates: constraint \\
Context: Of the eight students-George, Helen, Irving, Kyle, Lenore, Nina, Olivia, and Robert-in a seminar, exactly six will give individual oral reports during three consecutive days-Monday, Tuesday, and Wednesday. Exactly two reports will be given each day-one in the morning and one in the afternoon-according to the following conditions. \\
Constraint Rule: If Nina gives a report, then on the next day Helen and Irving must both give reports, unless Nina's report is given on Wednesday. \\
Symbolic Rule: \\  
- constraint(Helen, Irving, Tuesday) :- constraint(Nina, Monday) \\
- constraint(Helen, Irving, Wednesday) :- constraint(Nina, Tuesday)  \\

\#\#\# Here's what you need to do. \\
\textbf{Schema Objects:} \{objects\}  \\
\textbf{Schema Predicates:} \{predicates\}\\
\textbf{Context:} \{context\} \\
\textbf{Constraint Rule:} \{rule\} \\
\textbf{Symbolic Rule:}
\end{tcolorbox}
}
\caption{The prompt for rule initialization in AR-LSAT.}
\label{prompt_rule_initialization_ar_lsat}
\end{table*}

\begin{table*}
\begin{tcolorbox}[colback=blue!5!white,colframe=black!55!black,width=0.98\textwidth,title={Prompt for Rule Implementation (AR-LSAT)}]
\small
\textbf{System:} You are an expert in logical reasoning. You will receive a context including background information followed by a list of constraint rules, and a specific query with five candidate options (A, B, C, D, E). Your task is to accurately select the answer that satisfies the provided rule.\\

\textbf{User:} At each turn, you will be provided a context background, a constraint rule and a list of relevant facts. \\
Please on a new line starting with "Rule Implementation:" to implement the rule based on the facts to analyze there is a conflict between them. If no conflict, proceed to deduce new potential facts. \\
Then predict "Yes" or "No" to judge whether there is a conflict between the rule and facts, in a new line starting with "Judgement:". \\
If the judgement is No, proceed on a new line starting with "New fact:" to outline the new inferred fact in both natural language form and its corresponding symbolic format as "predicate(X, Y, ...)" within "[" and "]". \\
Please try to use the objects and predicates in the provided schema to describe symbolic facts. If the schema does not contain corresponding elements, generate the symbolic fact directly from its natural language form. Please always use one predicate, i.e., assign.\\

\#\#\# Examples: \\
Schema Objects: Monday, Tuesday, Wednesday, morning, Kyle, Lenore, Helen, George, Nina, Irving, Robert \\
Schema Predicates: assign \\
Context: Of the eight students-George, Helen, Irving, Kyle, Lenore, Nina, Olivia, and Robert-in a seminar, exactly six will give individual oral reports during three consecutive days-Monday, Tuesday, and Wednesday. Exactly two reports will be given each day-one in the morning and one in the afternoon-according to the following conditions. \\
Rule: Tuesday is the only day on which George can give a report. \\
Query: If Kyle and Lenore do not give reports, then the morning reports on Monday, Tuesday, and Wednesday, respectively, could be given by \\
Fact List: \\
- B) Irving, Robert, and Helen \\
Rule Implementation: According to the rule and the fact Robert give report on Tuesday morning, there is no conflict and we can infer George give a report on Tuesday afternoon. \\
Judgement: No. \\
New fact: George give a report on Tuesday afternoon. [assign(George, Tuesday, afternoon)] \\
------ \\
Schema Objects: Monday, Tuesday, Wednesday, morning, afternoon, Kyle, Lenore, Helen, George, Nina, Irving, Robert \\
Schema Predicates: assign \\
Context: Of the eight students-George, Helen, Irving, Kyle, Lenore, Nina, Olivia, and Robert-in a seminar, exactly six will give individual oral reports during three consecutive days-Monday, Tuesday, and Wednesday. Exactly two reports will be given each day-one in the morning and one in the afternoon-according to the following conditions. \\
Rule: Neither Olivia nor Robert can give an afternoon report. \\
Query: If Kyle and Lenore do not give reports, then the morning reports on Monday, Tuesday, and Wednesday, respectively, could be given by \\
Fact List: \\
- B) Irving, Robert, and Helen \\
- George give a report on Tuesday afternoon. \\
Rule Implementation: According to the rule, and the facts Irving, Robert, and Helen all give report on morning, there is a conflict that can not give a report on the morning. \\
Judgement: Yes. \\

\#\#\# Here's what you need to do. \\
\textbf{Schema Objects:} \{objects\} \\
\textbf{Schema Predicates:} \{predicates\}\\
\textbf{Context:} \{context\}\\
\textbf{Rule:} \{rule\} \\
\textbf{Query:} \{query\}\\
\textbf{Fact List:} \{facts\} \\
\textbf{Rule Implementation:}
\end{tcolorbox}
\caption{The prompt for LLM-based rule implementation in AR-LSAT.}
\label{prompt_rule_implementation_ar_lsat}
\end{table*}

\begin{table*}
\centering
\resizebox{0.98\textwidth}{!}{
\begin{tcolorbox}[colback=blue!5!white,colframe=black!55!black,width=0.98\textwidth,title={Prompt for State Fact Initialization (Boxes)}]
\small
Please list the symbolic form of the explicitly provided fact in the context. \\
Format the symbolic fact in Prolog notation as "predicate(X, Y, ...)" where X, Y, ... are the arguments of the predicate verb. \\
Please try to use the objects and predicates in the provided schema to describe symbolic facts. If the schema is null or does not contain corresponding elements to describe the context, generate the symbolic fact directly from its natural language form. \\

\#\#\# Examples: \\
Schema Objects: null \\
Schema Predicates: contains, move\_from\_to, remove\_from, put\_into \\
Context: Box 0 contains the rose. \\
Fact: contains(Box 0, the rose) \\
------ \\
Schema Objects: Box 0, the rose \\
Schema Predicates: contains, move\_from\_to, remove\_from, put\_into \\
Context: Box 4 contains the bread and the radio and the tape. \\
Fact: contains(Box 4, the bread, the radio, the tape) \\
------ \\
Schema Objects: Box 0, the rose \\
Schema Predicates: contains, move\_from\_to, remove\_from, put\_into \\
Context: Box 1 contains nothing. \\
Fact: contains(Box 1, nothing) \\

\#\#\# Here's what you need to do. \\
\textbf{Schema Objects:} \{objects\}  \\
\textbf{Schema Predicates:} contains, move\_from\_to, remove\_from, put\_into \\
\textbf{Context:} \{context\} \\
\textbf{Fact:}
\end{tcolorbox}
}
\caption{The prompt for state fact initialization in Boxes.}
\label{prompt_fact_initialization_boxes}
\end{table*}

\begin{table*}
\centering
\resizebox{0.98\textwidth}{!}{
\begin{tcolorbox}[colback=blue!5!white,colframe=black!55!black,width=0.98\textwidth,title={Prompt for Operation Fact Initialization (Boxes)}]
\small
Please list the symbolic form of the explicitly provided fact in the context. \\
Format the symbolic fact in Prolog notation as "predicate(X, Y, ...)" where X, Y, ... are the arguments of the predicate. \\
Please try to use the objects and predicates in the provided schema to describe symbolic facts. If the schema is null or does not contain corresponding elements to describe the context, generate the symbolic fact directly from its natural language form. \\

\#\#\# Examples: \\
Schema Objects: null \\
Schema Predicates: contains, move\_from\_to, remove\_from, put\_into \\
Context: Put the shoe into Box 0. \\
Fact: put\_into(the shoe, Box 0) \\
------ \\
Schema Objects: null \\
Schema Predicates: contains, move\_from\_to, remove\_from, put\_into \\
Context: Remove the radio and the tape from Box 4. \\
Fact: remove\_from(the radio, the tape, Box 4) \\
------ \\
Schema Objects: Box 0, the rose, the bread, the radio, the tape \\
Schema Predicates: contains, move\_from\_to, remove\_from, put\_into \\
Context: Move the contents of Box 3 to Box 1. \\
Fact: move\_from\_to(the contents, Box 3, Box 1) \\

\#\#\# Here's what you need to do. \\
\textbf{Schema Objects:} \{objects\}  \\
\textbf{Schema Predicates:} contains, move\_from\_to, remove\_from, put\_into \\
\textbf{Context:} \{context\} \\
\textbf{Fact:}
\end{tcolorbox}
}
\caption{The prompt for operation fact initialization in Boxes.}
\label{prompt_op_fact_initialization_boxes}
\end{table*}

\begin{table*}
\begin{tcolorbox}[colback=blue!5!white,colframe=black!55!black,width=0.98\textwidth,title={Prompt for Rule Implementation (Boxes)}]
\small
\textbf{System:} You are an expert in logical reasoning. You will receive a context including a list of state facts and operational facts, a list of rules and a specific query. Your task is to answer this query following the provided rule.\\

\textbf{User:} At each turn, you will be provided a list of state facts and an operational fact, and a logical rule. \\
Please on a new line starting with "Rule Implementation:" to implement the rule based on the facts to infer new state facts after the operation. \\
Then output "New facts:" in a new line, and each new inferred fact in both natural language form and its corresponding symbolic format on separate lines under the header "New facts:".  \\
Each line must cover all contents about a distinct Box. For example, the first is about Box 1, then the second line should not describe Box 1. \\
Format each fact in natural language as "Box X contains Y." where X is the box number and Y are the specific items instead of general "contents" in the box. 
Format each symbolic fact in Prolog notation as "predicate(X, Y, ...)" where X, Y, ... are the arguments of the predicate, and the predicate should be "contains". 
Please try to use the objects and predicates in the provided schema to describe symbolic facts. If the schema does not contain corresponding elements, generate the symbolic fact directly from its natural language form.  \\

\#\#\# Examples: \\
Schema Objects: Box 0, the rose, the bread, the radio, the tape  \\
Schema Predicates: contains, move\_from\_to, remove\_from, put\_into \\
State Facts: Box 1 contains the rose. Box 2 contains the letter. \\
Operational Fact: Move the contents from Box 2 to Box 1. \\
Rule: If move the contents X from Box A to Box B, then X are not in Box A and X are in Box B. \\
Rule Implementation: Based on the rule, after the moving operation, we can infer that Box 1 contains the rose and the letter, and Box 2 contains nothing. \\
New facts: \\
Box 1 contains the rose and the letter. [contains(Box 1, the rose, the letter)] \\
Box 2 contains nothing. [contains(Box 2, nothing)] \\
------ \\
Schema Objects: Box 0, Box 1, Box 2, the rose, the bread, the radio, the tape, the letter, the book, nothing \\
Schema Predicates: contains, move\_from\_to, remove\_from, put\_into \\
State Facts: Box 2 contains the letter and the book. \\
Operational Fact: Remove the letter from Box 2. \\
Rule: If remove the contents X from Box A, then X are not in Box A. \\
Rule Implementation: Based on the rule, after the removing operation, we can infer that Box 2 contains the book. \\
New facts:  \\
Box 2 contains the book. [contains(Box 2, the book)] \\

\#\#\# Here's what you need to do. \\
\textbf{Schema Objects:} \{objects\} \\
\textbf{Schema Predicates:} \{predicates\}\\
\textbf{State Facts:} \{state facts\}\\
\textbf{Operational Fact:} \{op facts\}\\
\textbf{Rule:} \{rule\} \\
\textbf{Rule Implementation:}
\end{tcolorbox}
\caption{The prompt for LLM-based rule implementation in Boxes.}
\label{prompt_rule_implementation_boxes}
\end{table*}

\end{document}